\documentclass[letterpaper, 10 pt, conference]{ieeeconf}  

\IEEEoverridecommandlockouts 

\usepackage{amsmath,amsfonts}
\usepackage{graphicx}
\usepackage{textcomp}
\usepackage{comment}

\usepackage{amsthm}
\usepackage[utf8]{inputenc}
\usepackage[T1]{fontenc}
\usepackage{enumerate}
\usepackage{soul}
\usepackage[dvipsnames,table]{xcolor} 
\usepackage{hyperref}
\usepackage{siunitx}
\usepackage{multirow}
\usepackage{subfig}
\usepackage{booktabs}
\usepackage[ruled,vlined]{algorithm2e}
\usepackage{algpseudocode}
\DontPrintSemicolon

\usepackage{amssymb}    
\usepackage{mathtools}  
\usepackage{bm}         
\usepackage{accents}
\usepackage{caption}
\usepackage{colortbl}  
\usepackage{array}     
\usepackage{adjustbox} 
\usepackage{wasysym}  
\usepackage{pifont}
\usepackage{stfloats}
\usepackage{url}
\usepackage{verbatim}
\usepackage{makecell}
\usepackage{tikz}
\usepackage{color}  
\usepackage{lipsum} 
\usepackage{float}

\usepackage[style=ieee,backend=biber,sortcites=true]{biblatex}

\usepackage{fontawesome5}

\renewcommand*{\multicitedelim}{\addcomma\space}
\DeclareCiteCommand{\cite}[\mkbibbrackets]
  {\usebibmacro{prenote}}
  {\usebibmacro{citeindex}\printfield{labelnumber}}
  {\multicitedelim}
  {\usebibmacro{postnote}}
  
\AtBeginBibliography{\scriptsize}  
\definecolor{blue}{HTML}{4E95D9}
\definecolor{gold}{HTML}{E5AE1C}
\definecolor{safe}{HTML}{4169E1}
\definecolor{unsafe}{HTML}{C41230}

\newcommand{\ubar}[1]{\underaccent{\bar}{#1}}  

\DeclareMathOperator*{\argmin}{argmin}

\usetikzlibrary{shapes, arrows, arrows.meta, positioning, shapes.geometric, decorations.pathreplacing, calc, patterns}

\definecolor{dangerRed}{HTML}{C00000}
\definecolor{dangerFill}{HTML}{FBE5E5}
\definecolor{nomGray}{HTML}{7F7F7F}

\definecolor{llampc}{HTML}{9B5EA9}
\newcommand{\llampc}{\textbf{{\textcolor{llampc}{LLA-MPC}}}}

\newcommand\copyrighttext{%
    \normalfont\footnotesize  This work has been accepted in the Sim2Real and Classical Control: From Rigorous Theory to Data-Driven Robotics Workshop at the 2026 IEEE/RSJ International Conference on Intelligent Robots and Systems (IROS 2026). Link:
 \href{https://sim2realgap.github.io/sim2real-and-control-workshop-iros2026/}{\textcolor{llampc}{https://sim2realgap.github.io/sim2real-and-control-workshop-iros2026/}}}%
\newcommand\copyrightnotice{%
\begin{tikzpicture}[remember picture,overlay]
\node[anchor=south,yshift=10pt] at (current page.south) {\fbox{\parbox{\dimexpr\textwidth-\fboxsep-\fboxrule\relax}{\copyrighttext}}};
\end{tikzpicture}%
}

\title{\bf
\llampc\ on Embedded Hardware: \\ Rapid Adaptive Control with Thousands of Parallel Models}
\author{Henry Z. Liao$^\bigstar$ \quad Maitham F.~AL-Sunni$^\bigstar$\quad  John M. Dolan%
\thanks{$^\bigstar$ These authors contributed equally to the work.}%
\thanks{H. Z. Liao and J. M. Dolan are with the Robotics Institute, Carnegie Mellon University, Pittsburgh, PA, USA. Email: {\tt \{hzl,jdolan\}@andrew.cmu.edu}}%
\thanks{M. F. AL-Sunni is with the Department of Electrical \& Computer Engineering, Carnegie Mellon University, Pittsburgh, PA, USA. Email: {\tt  malsunni@andrew.cmu.edu}}%
}

\begin{document}

\maketitle

\begin{abstract}
We present a generalized implementation of \textcolor{llampc}{Look-Back and Look-Ahead Adaptive
Model Predictive Control (LLA-MPC)}, a learning-free framework for real-time,
rapid adaptive system identification and control. The original formulation
was demonstrated only in simulation, for autonomous racing with a fixed model
structure. Our implementation has modular dynamics and integrators, and we
validate it on the F1TENTH platform under constrained computation and noisy
state estimation. The system identifies tire parameters online by evaluating
thousands of candidate models in real time on an embedded computer.
Experiments with low-friction tires across changing surfaces show that
\llampc\ completes high-speed tracking tasks where a fixed nominal model
fails. Code, videos, and our related work are available at
\href{https://lla-control.github.io}{\textcolor{llampc}{lla-control.github.io}}.%
\copyrightnotice
\end{abstract}

\section{Introduction}
\label{sec:introduction}

Model-based control for robots degrades when dynamics change rapidly during operation. Car tires wear, surfaces change, and payloads shift. Recovering performance requires identification that reacts within a fraction of a second. Look-back and Look-ahead Adaptive Model Predictive Control (\llampc) addresses this without learning \cite{alsunni2025llampc}. The framework holds a large bank of candidate models with varied parameter values and evaluates their one-step predictions in a single batched computation. At every step, the candidate with the best performance over a short {\it look-back} window becomes the prediction model of a receding-horizon controller.

\llampc\ runs an ensemble of physics simulators onboard, and the look-back selection
re-aligns it with reality. The original paper demonstrated the
framework only in simulation, and its vehicle model did not survive the
transfer to hardware. This paper presents a generalized implementation that
bridges this gap. We deploy it on the F1TENTH platform \cite{f1tenth2020},
with a new single-track dynamics module and Fiala brush tire forces, on
embedded computation, with onboard state estimation in
extreme driving conditions.

\subsection{Related Work}
Adaptive control with multiple models is a classical idea. Multiple-model adaptive control switches among pre-designed controllers paired with parallel identification models \cite{narendra1997adaptive}, multiple-model predictive control blends a bank of linear models by residual weights \cite{aufderheide2003extension}, and adaptive MPC estimates parameters continuously inside the controller \cite{adetola2009adaptive}. \llampc\ instead keeps a single receding-horizon optimizer and selects among thousands of nonlinear physics models by batched prediction error \cite{alsunni2025llampc}. A recent method also selects models online from residual errors, but its library spans models of different fidelity and cost to balance accuracy against computation \cite{cancelliere2026ado}, whereas the \llampc\ bank varies one model's parameters to follow changing dynamics.

Online adaptation methods for racing usually learn a model or a correction, either beforehand or during operation. They update a Gaussian process residual
during the race \cite{kabzan2019learning}, learn tire residuals after a warm-up lap \cite{10610848}, calibrate a tire model from seconds of driving
\cite{dikici2025learning}, or pair online estimation with offline-trained components \cite{nguyen2026ad}. Meta-learned and pretrained models adapt online only after offline training \cite{levy2025metalearning,
kalaria2025agile, xiao2025anycar,mungiello2026vehicle}. Robust formulations skip identification and hedge against a fixed model set \cite{abraham2020model,lew2025riskaverse}, and domain randomization trains one policy across randomized simulation parameters
\cite{tobin2017domain, peng2018sim}. \llampc\ holds the same kind of
ensemble as a bank of simulators and identifies at run time which one
matches reality. It requires no training, no warm-up data, and no fitting
step, so it adapts within a look-back window of a change rather than after
a data collection phase, and its identified parameters remain physical
quantities.

\subsection{Contributions}
The contributions of this work are threefold.
\begin{enumerate}
    \item A generalized, open-source implementation of \llampc\ with modular dynamics, integrators, and parallelization backends.
    \item Hardware validation of \llampc\ under constrained computation and noisy state estimation, showing improved performance over a fixed nominal model.
    \item Demonstrations that show the robustness of \llampc\ across tasks, noise levels, and environmental conditions.
\end{enumerate}

\section{Methodology}
\label{sec:methodology}

Consider the discrete-time dynamical system
\begin{equation}
    {\mathbf{x}_{t+1} = f(\mathbf{x}_t, \mathbf{u}_t)},
\end{equation}
where $\mathbf{x} \in \mathbb{R}^n$ is the state, $\mathbf{u} \in \mathbb{R}^m$ is the control, and $f:\mathbb{R}^n \times \mathbb{R}^m\rightarrow \mathbb{R}^n $ is the unknown dynamics. Model-based methods require a model of $f$. We consider a parametric candidate model $\widetilde f(\mathbf{x}, \mathbf{u};\,\boldsymbol{\theta})$, with parameters $\boldsymbol{\theta} \in \mathbb{R}^{p}$, whose structure is chosen to approximate $f$. At time step $t$, given a recorded trajectory of the last $W$ transitions, $\mathcal{D}_t = \{\mathbf{x}_{t-W}, \mathbf{u}_{t-W}, \ \dots \mathbf{x}_t\}$, generated by the true dynamics $f$, identification can be posed as solving for the optimal parameters
\begin{equation}
\label{eq:sysid}
  \boldsymbol{\theta}^\star = \argmin_{\boldsymbol{\theta}} \; \mathcal{E}\big(\boldsymbol{\theta}; \mathcal{D}_t\big),
\end{equation}
where the error metric $\mathcal{E}$ evaluates the accuracy of the candidate
model $\widetilde f(\cdot,\cdot\,;\boldsymbol{\theta})$ in predicting the
recorded trajectory. Since $\mathcal{D}_t$ contains only the last $W$
transitions, the identification in \eqref{eq:sysid} is local in time and,
because the trajectory is continuous, local in the region of the state and
control space the system is visiting. The goal is the model that best
describes the dynamics locally, in time, where it operates, and when
the true dynamics change, the minimizer of \eqref{eq:sysid} moves with them.
Classical identification estimates $f$ offline from large datasets and holds
it fixed. Recursive estimators such as least-squares and gradient updates
adapt the model online, but they need model derivatives or linearity in the
parameters, and their single running estimate must travel across the
parameter space when the dynamics change abruptly. \llampc\ instead solves
\eqref{eq:sysid} by direct evaluation over a fixed set of candidates. It
needs only forward simulation, holds its selection when the data is
uninformative, and can jump to a distant candidate as soon as the window
supports one. 

\subsection{Look-Back Adaptive System Identification}
We formulate a bank of candidate models by varying $\boldsymbol{\theta}$ uniformly across the parameter space $\ubar{\boldsymbol{\theta}} \leq \boldsymbol{\theta} \leq \bar{\boldsymbol{\theta}}$, where $\ubar{\boldsymbol{\theta}}$ and $\bar{\boldsymbol{\theta}}$ are vectors that represent the lower and upper bounds of the parameter space, respectively. This results in a bank of $K$ candidate models:
\begin{align}
    \mathcal{F}
    :=
    \left\{
    \widetilde{f}(\cdot,\cdot\,;\bm{\theta}^{i})
    \right\}_{i=1}^{K}.
\end{align}
We continue by taking inspiration from the Path Integral formulations for control \cite{williams2016mppi}, \cite{williams2018mppi} that enable parallelization by the fact that dynamics only vary \textit{numerically} with respect to states and time, but not functionally.  \llampc\ adds the insight that similarly, dynamics only vary numerically with respect to parameters. Since the arithmetic functions of $\mathcal{F}$ hold constant across all models, the rollouts for the bank of models are easily parallelizable.

At each timestep $t$, \llampc\ calculates the state predictions for all models in parallel based on current state $\mathbf{x}_t$ and control $\mathbf{u}_t$ to be applied, as follows
\begin{align}
        \mathcal{F}_{\mathbf{x}_t,\mathbf{u}_t}
    =
    \left\{
        \widetilde{f}\!\left(
            \mathbf{x}_t,\mathbf{u}_t;\bm{\theta}^{i}
        \right)
    \right\}_{i=1}^{K}
    =
    \left\{
        \widetilde{\mathbf{x}}_{t+1}^{i}
    \right\}_{i=1}^{K}.
\end{align}
Here $\mathcal{F}_{\mathbf{x},\mathbf{u}}$ denotes the bank evaluated at $(\mathbf{x},\mathbf{u})$. At the next timestep $t+1$, for each model $i$ in the bank, the residual $\mathbf{e}^{i}_{t+1} = \widetilde{\mathbf{x}}^{i}_{t+1} - \mathbf{x}_{t+1}$ can be calculated based on the prediction of the $i$-th model in the bank $\widetilde{\mathbf{x}}^{i}_{t+1}$ and the true state $\mathbf{x}_{t+1}$, at which point \llampc\ uses a quadratic norm to evaluate the error between the  predictions and sensor reading as
\begin{align}
\label{eq:llanorm}
     ||\mathbf{e}^i_{t+1}||_S^2 = (\mathbf{e}^i_{t+1})^TS(\mathbf{e}^i_{t+1}), \quad \forall i \in \{1, 2, \ \dots K\},
\end{align}
where $S$ is a diagonal matrix specifying \llampc\ sensitivity to states.

\llampc\ formulates the identification \eqref{eq:sysid} as the summation of the norm of the residual \eqref{eq:llanorm} over the last $W$ timesteps
\begin{align}
\label{eq:llaopt}
\mathcal{E}_t^i = \mathcal{E}\big(\boldsymbol{\theta}^i; \mathcal{D}_t\big) = \sum_{k=t-W+1}^{t} \lVert\mathbf{e}^i_{k}\rVert_{S}^{2}
\end{align}
$\forall i \in \{1, 2, \ \dots K\}$, and it selects the model $\boldsymbol{\theta}^{i^\star_t}$ that minimizes the accumulated quadratic error 
\begin{align}
i^\star_t =\underset{i}\argmin \;\mathcal{E}^i_t.
\end{align}
\llampc\ maintains a parallelized sliding window \cite{slidingwindow1985} for fast tracking and updating of the summation in \eqref{eq:llaopt}. The last $W$ residual norms $\{||\mathbf{e}^i_{t-W+1}||_S^2, ||\mathbf{e}^i_{t-W+2}||_S^2, \ \dots ||\mathbf{e}^i_{t}||_S^2\}$ are cached for all $K$ models. In conjunction with \eqref{eq:llanorm}, this means the running cost can be updated recursively as
\begin{equation}
\label{eq:slidingwindow}
\mathcal{E}^i_{t+1}  = \mathcal{E}^i_t+ 
||\mathbf{e}^i_{t+1}||_S^2 - ||\mathbf{e}^i_{t-W+1}||_S^2,
\end{equation}
$\forall i \in \{1, 2, \ \dots K\}$. The cache is implemented as a circular array for memory efficiency, where after the update \eqref{eq:slidingwindow}, $||\mathbf{e}^i_{t-W+1}||_S^2$ is overwritten by $||\mathbf{e}^i_{t+1}||_S^2$ for all $K$ models.

\subsection{Look-Ahead Model Predictive Control}

After identifying the optimal parameter set
$\boldsymbol{\theta}^{i_t^\star}$ at time $t$, we use the
corresponding best-fit model
$\widetilde{f}(\mathbf{x},\mathbf{u};
\boldsymbol{\theta}^{i_t^\star})$ for model predictive control.
Given the current state $\mathbf{x}_t$, we solve
\begin{align}
\label{eq:mpc}
\min_{\mathbf{u}_0,\ldots,\mathbf{u}_{H-1}}\quad
&\sum_{h=0}^{H-1}
\Big[
\ell(\mathbf{x}_h,\mathbf{x}_{\mathrm{ref},h},\mathbf{u}_h)
+\ell_\Delta(\mathbf{u}_h,\mathbf{u}_{h-1})
\Big] \\
&+\ell_T(\mathbf{x}_H,\mathbf{x}_{\mathrm{ref},H}) \nonumber \\
\mathrm{s.t.}\quad
&\mathbf{x}_0=\mathbf{x}_t, \nonumber \\
&\mathbf{x}_{h+1}
=\widetilde{f}(\mathbf{x}_h,\mathbf{u}_h;
\boldsymbol{\theta}^{i_t^\star}),
\quad h=0,\ldots,H-1, \nonumber \\
&\ubar{\mathbf{x}}
\leq \mathbf{x}_h
\leq \bar{\mathbf{x}},
\quad h=0,\ldots,H, \nonumber \\
&\ubar{\mathbf{u}}
\leq \mathbf{u}_h
\leq \bar{\mathbf{u}},
\quad h=0,\ldots,H-1, \nonumber \\
&\ubar{\mathbf{u}}_\Delta
\leq \mathbf{u}_h-\mathbf{u}_{h-1}
\leq \bar{\mathbf{u}}_\Delta,
\quad h=0,\ldots,H-1. \nonumber
\end{align}
Here, $\mathbf{u}_{-1}$ denotes the previously applied control input and is fixed during the optimization. The optimization horizon is $H$, and $\mathbf{x}_{\mathrm{ref},h}\in\mathbb{R}^n$, $h\in\{0,\ldots,H\}$, denotes the reference trajectory provided by the planner. The functions $\ell$, $\ell_\Delta$, and $\ell_T$ denote the running, control-rate, and terminal costs, respectively. The underlined and overlined quantities denote the corresponding lower and upper bounds.

\section{Experimental Validation: Extreme Driving}
\label{sec:application}
As a case study, we integrate a single-track dynamic bicycle model with tire-road interactions into the \llampc\ framework, and validate the efficacy of \llampc\ system identification and control with the F1TENTH platform \cite{f1tenth2020} across a wide range of extreme driving scenarios.

\subsection{System Dynamics}
\label{subsec:systemdynamics}

\subsubsection{Single-Track Vehicle Model} 
We model the vehicle dynamics where the state $\mathbf{x}$ is described as follows: $(x, y, \varphi)$ is the inertial pose, $(v_x, v_y, \omega)$ are the body-frame
longitudinal, lateral, and yaw rates, and $\omega_w$ is the driveline (wheel) speed. The controls $\mathbf{u}$ are the motor current $i$ and the road-wheel steering angle $\delta$. The dynamical system is formulated as
\begin{align}
\label{eq:rigid_body}
  {
  \begin{bmatrix}
    \dot{x}       \vphantom{v_x \cos\varphi - v_y \sin\varphi} \\
    \dot{y}       \vphantom{v_x \sin\varphi + v_y \cos\varphi} \\
    \dot{\varphi} \vphantom{\omega} \\
    \dot{v}_x     \vphantom{\frac{1}{m}\left( F_{\mathbf{r}x} + F_{\mathbf{f}x}\cos\delta - F_{\mathbf{f}y}\sin\delta \right) + v_y\,\omega} \\
    \dot{v}_y     \vphantom{\frac{1}{m}\left( F_{\mathbf{r}y} + F_{\mathbf{f}y}\cos\delta + F_{\mathbf{f}x}\sin\delta \right) - v_x\,\omega} \\
    \dot{\omega}  \vphantom{\frac{1}{I_z}\left( l_\mathbf{f}\left(F_{\mathbf{f}y}\cos\delta + F_{\mathbf{f}x}\sin\delta\right) - l_\mathbf{r} F_{\mathbf{r}y} \right)} \\
    \dot{\omega}_w\vphantom{\frac{\tau_d - \left(F_{\mathbf{r}x} + F_{\mathbf{f}x}\right) r_w}{I_w + I_m}}
  \end{bmatrix}}
  =
  {
  \begin{bmatrix}
    v_x \cos\varphi - v_y \sin\varphi \\
    v_x \sin\varphi + v_y \cos\varphi \\
    \omega \\
    \frac{1}{m}\left( F_{\mathbf{r}x} + F_{\mathbf{f}x}\cos\delta - F_{\mathbf{f}y}\sin\delta \right) + v_y\,\omega \\
    \frac{1}{m}\left( F_{\mathbf{r}y} + F_{\mathbf{f}y}\cos\delta + F_{\mathbf{f}x}\sin\delta \right) - v_x\,\omega \\
    \frac{1}{I_z}\left( l_\mathbf{f}\left(F_{\mathbf{f}y}\cos\delta + F_{\mathbf{f}x}\sin\delta\right) - l_\mathbf{r} F_{\mathbf{r}y} \right) \\
    \frac{\tau_d - \left(F_{\mathbf{r}x} + F_{\mathbf{f}x}\right) r_w}{I_w + I_m}
  \end{bmatrix}}.
\end{align}
The discrete dynamics $\widetilde f$ is obtained by integrating the continuous \eqref{eq:rigid_body} over the control period $\Delta t$.

We identified the following static parameters: vehicle mass $m$, yaw inertia $I_z$, the distances from the center of mass to the front and rear axles $l_\mathbf{f}$ and $l_\mathbf{r}$, wheel radius $r_w$ , driveline inertia $I_w$, and the approximate rotor inertia reflected in wheel rotation $I_m$. The drive torque $\tau_d$ follows the standard relation for Permanent Magnet Synchronous Motor (PMSM) control, given by
$\tau_d = \tfrac{3}{2}\, G\, n_p\, \lambda_m\, i$,
with gear ratio $G$, pole pairs $n_p$, and rotor flux linkage $\lambda_m$.

\subsubsection{Fiala Brush Tire Model}
To model tire-ground interactions, we use the Fiala Tire Model with a friction circle constraint inspired by prior works detailing its efficacy for on-hardware control \cite{lew2025riskaverse}, \cite{zhang2018fiala}. The model uses a ratio to calculate lateral friction limits based on normal forces:
\begin{alignat}{2}
  F_{zj} = \frac{m g\, l_{j'}}{(l_\mathbf{f}+l_\mathbf{r})}, \
  \bar{F}_{j} = \mu_j F_{zj},
  \ j \in \{ \mathbf{f}, \mathbf{r}\},
\end{alignat}
where $j$ represents the front and rear axles, $j'$ represents the opposite axle to $j$, ${F}_{zj}$ represents the normal force load per axle, and  $\bar{F}_{j}$ represents the per-axle friction limits.
 
The front and rear drive wheels on the F1TENTH are coupled together, but only the front tires steer, resulting in the following slip angles $\alpha$ and slip ratios $\kappa$:
\begin{alignat}{2}
  \alpha_\mathbf{f} &= \delta - \tan^{-1}\left(\frac{v_y + l_\mathbf{f} \omega}{v_x}\right),
  &\quad
   \kappa_\mathbf{f} &= \frac{r_w \omega_w - v_{x,\mathbf{f}}}{v_{x,\mathbf{f}}},\\
  \alpha_\mathbf{r} &= \tan^{-1}\left(\frac{l_\mathbf{r} \omega - v_y}{v_x}\right),
  &\quad
  \kappa_\mathbf{r} &= \frac{r_w \omega_w - v_x}{v_x},
\end{alignat}
where $v_{x,\mathbf{f}}$ is the longitudinal velocity measured in the frame of the front steer wheels.

Lateral and longitudinal slip are combined into a single resultant slip magnitude per axle $\sigma_j = \sqrt{\tan^2\!\alpha_j + \kappa_j^2}$.
The Fiala brush model gives the friction as a cubic in $\sigma_j$ up to full sliding, and saturates at the friction limit thereafter:
\begin{align}
\label{eq:fiala}
  F_j =
  &\begin{cases}
    C_j \sigma_j
    - \dfrac{\left(C_j \sigma_j\right)^2}{3 \bar{F}_j}
    + \dfrac{\left(C_j \sigma_j\right)^3}{27 \bar{F}_j^{\,2}},
    & \sigma_j < \sigma_j^{\max}, \\
    \bar{F}_j, & \sigma_j \geq \sigma_j^{\max},
  \end{cases}
\end{align}
where $  \sigma_j^{\max} = \frac{3 \bar{F}_j}{C_j}$.

The calculated friction $F_j$ is distributed between the lateral and longitudinal directions in proportion to the slip components $ F_{jy} = F_j\, \frac{\tan\alpha_j}{\sigma_j}, \
  F_{jx} = F_j\, \frac{\kappa_j}{\sigma_j}$, which enforces the friction circle $\sqrt{F_{jx}^2 + F_{jy}^2} \le \bar{F}_j$ implicitly.

The standard parameters of the Fiala Tire model \\
$\boldsymbol{\theta} = \begin{bmatrix} C_\mathbf{f} & C_\mathbf{r} & \mu_\mathbf{f} & \mu_\mathbf{r}\end{bmatrix}^{\!\top} \in \mathbb{R}^4$
are the ones that we consider in our \llampc\ implementation.

\subsection{Experimental Setup}
We deploy the dynamics from \autoref{subsec:systemdynamics} in our generalized \llampc\ framework on the F1TENTH platform with plastic tires that slip at extremely low wheel speeds.

We set up 2 main experimental configurations to evaluate the \llampc\ method under varying computation and state-estimation conditions. The first was with pose estimated by motion capture cameras at 180 Hz, and with all \llampc\ computation running off-board on an Intel 258V CPU laptop at up to 40 Hz. The second was with pose estimated by an onboard particle filter at 40 Hz, with all computation run on the CPU of a Jetson Orin Nano. In both cases, 5000 \llampc\ models were integrated by a parallel  4th-order Runge-Kutta (RK4) method that we implemented in {\tt JAX} \cite{JAX}. Using the GPU of the Jetson Orin Nano allowed \llampc\ to roll out at least 15000 models simultaneously at 40 Hz alongside all other necessary computation. The MPC was evaluated on multiple path-tracking tasks, with reference paths generated by a planner that interpolated predefined waypoints. Each task was attempted first by the MPC with a nominal model (identified by a grid search based on driving data with rubber tires at normal speeds) and then with \llampc. Performance was measured by the ability to complete the tracking task without catastrophic failure, which manifested as either significant deviation from the reference path causing a crash with the environment or a complete stop without the ability to continue the tracking task. The performance of system identification was evaluated by the $H$-step prediction error, defined as the deviation of the identified model's one-step prediction from the measured trajectory over the next $H$ timesteps, given by
\begin{equation}
\label{eq:hstep_error_avg}
\mathcal{P}_t = \sum_{h=t}^{t+H-1} \big\lVert \widetilde{f}\big(\mathbf{x}_{h}, \mathbf{u}_{h};
\boldsymbol{\theta}^{i^\star_t}\big) - \mathbf{x}_{h+1} \big\rVert_{S}^{2}\ .
\end{equation}
The parameter space for the implementation of the Fiala tire model is given by the lower bound $\ubar{\boldsymbol{\theta}} = \begin{bmatrix} 175 & 150 & 0.1 & 0.1 \end{bmatrix}^{\!\top}$ and the upper bound $\bar{\boldsymbol{\theta}} = \begin{bmatrix} 325 & 300 & 1.1 & 1.1 \end{bmatrix}^{\!\top}$. The \llampc\ window size is $W=60$. The cost functions for our implementation of the MPC problem \eqref{eq:mpc} in {\tt acados} \cite{acados2021} are defined as $\ell(\mathbf{x}_h, \mathbf{x}_{\text{ref}, h}, \mathbf{u}_h) = ||\mathbf{x}_h-\mathbf{x}_{\text{ref}, h}||_Q^2 + ||\mathbf{u}_h||_R^2$, $\ell_\Delta(\mathbf{u}_h, \mathbf{u}_{h-1}) = ||\mathbf{u}_h-\mathbf{u}_{h-1}||_{R_\Delta}^2$,
$\ell_T(\mathbf{x}_H, \mathbf{x}_{\text{ref}, H}) = ||\mathbf{x}_H- \mathbf{x}_{\text{ref}, H}||_{Q_T}^2$, where $Q\in \mathbb{R}^{n\times n}$, $R\in \mathbb{R}^{m\times m}$, $R_\Delta \in \mathbb{R}^{m\times m}$, $Q_T\in \mathbb{R}^{n\times n}$ are diagonal matrices. The optimization horizon is $H=30$. Violations of state and control bounds are penalized by a slack cost \cite{boyd2004convex}.

\subsection{Experiment 1: Tracking}

\begin{figure}[th]
    \centering
    \vspace{-2mm}
    \includegraphics[width=.49\linewidth]{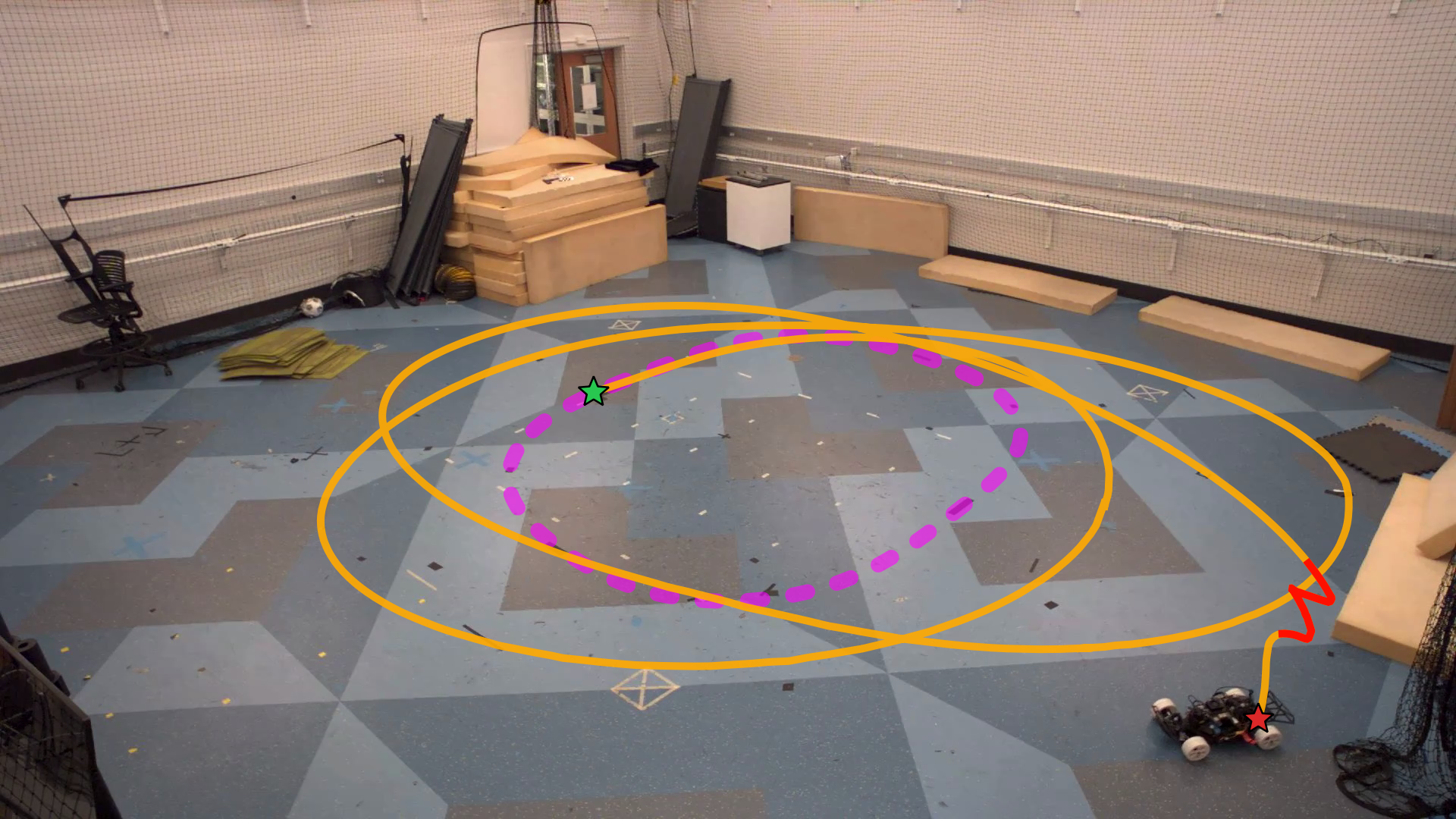}
    \includegraphics[width=.49\linewidth]{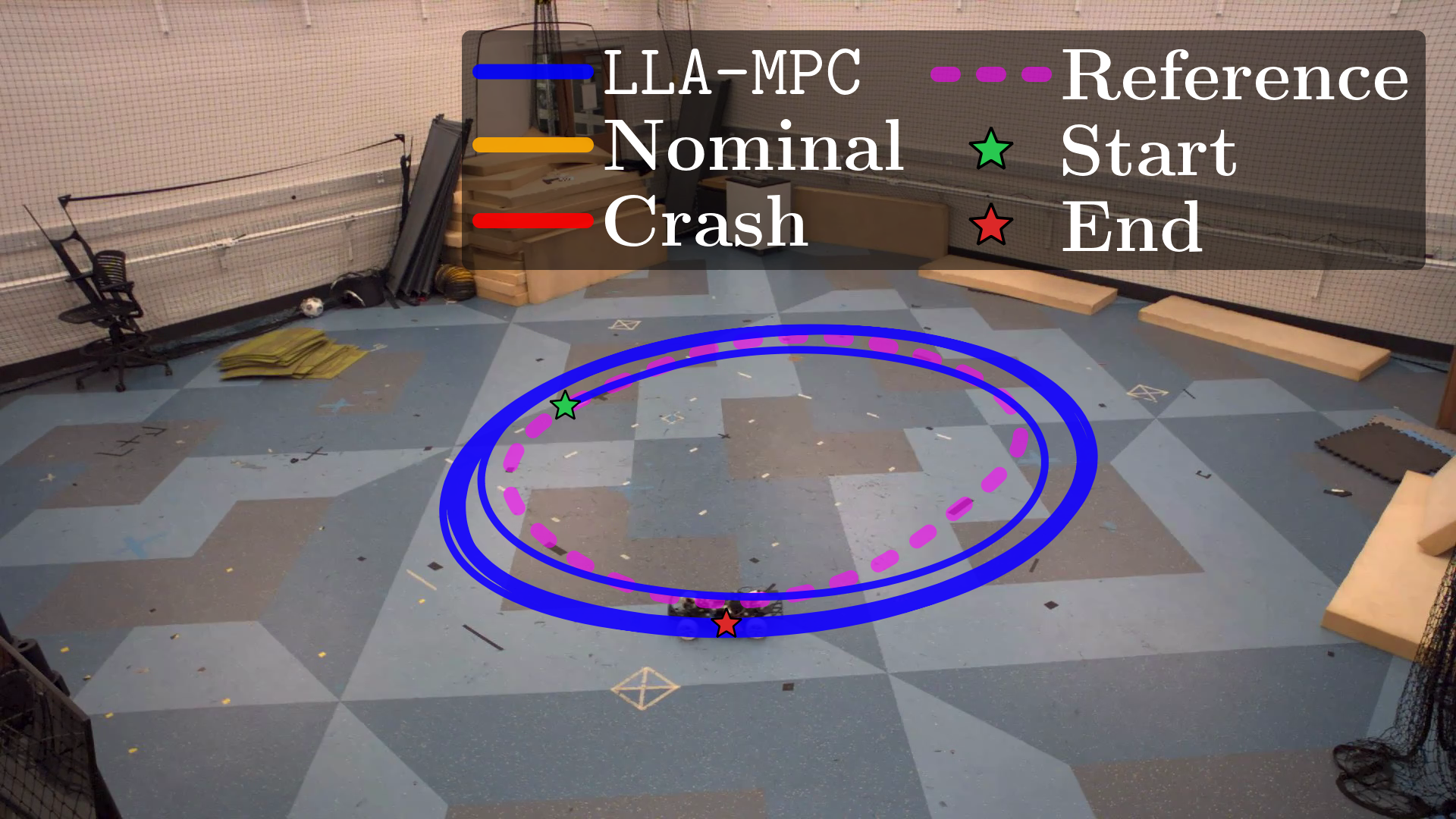}
    \caption{Nominal model tracking shown to the left. \llampc\ shown on the right.}
    \label{fig:drift_traj}
    \vspace{-6mm}
\end{figure}

\begin{figure}[th]
    \centering
    \vspace{-2mm}
    \includegraphics[width=.49\linewidth]{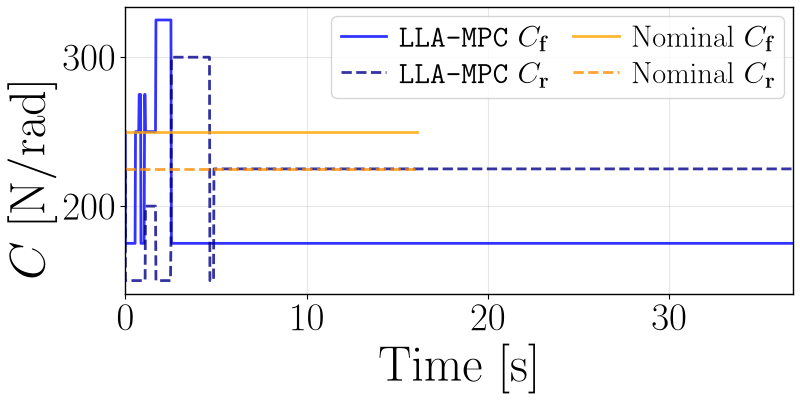}
    \includegraphics[width=.49\linewidth]{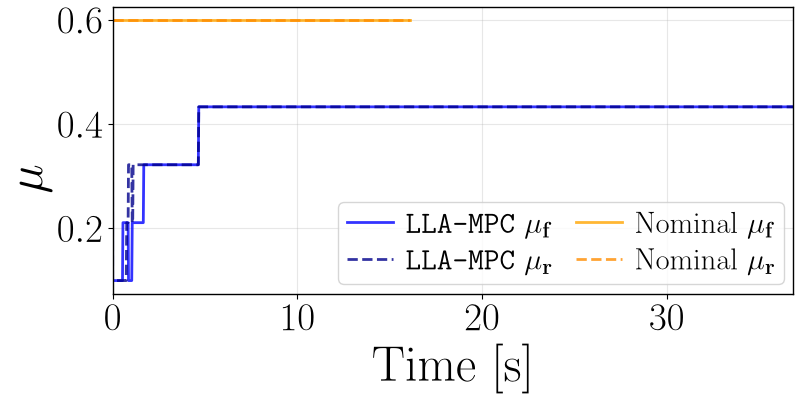}
    \includegraphics[width=.49\linewidth]{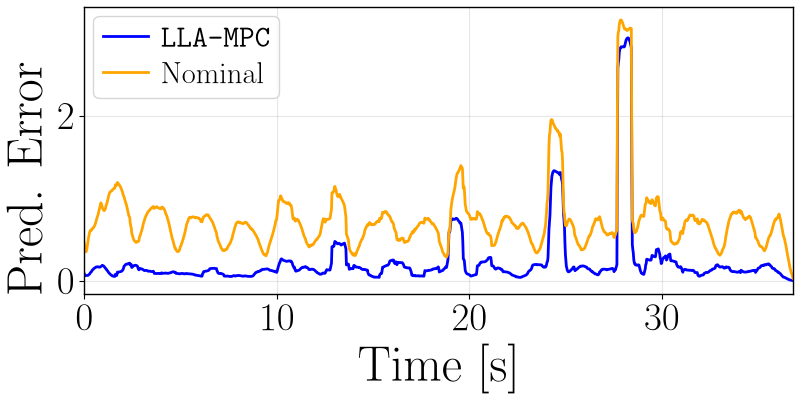}
    \includegraphics[width=.49\linewidth]{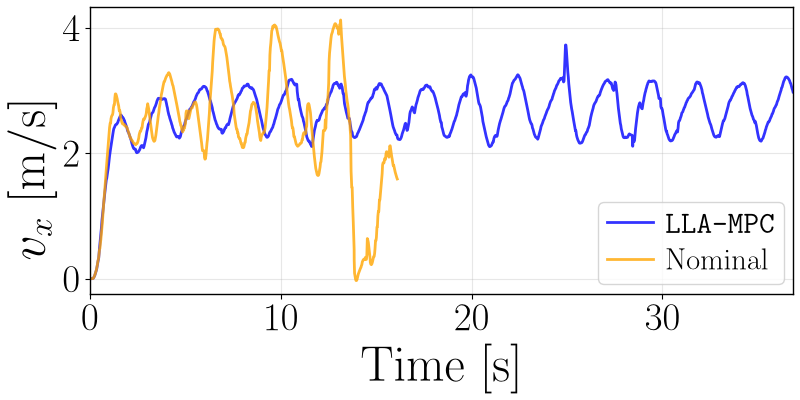}
    \caption{Model parameter estimation, longitudinal velocity, and $H$-step prediction error for run with \llampc.}
    \label{fig:drift_params}
    \vspace{-3mm}
\end{figure}

We first present a high-speed tracking task, where the robot tracks a trajectory at fast speed under low friction. \llampc\ is able to track the trajectory at the same average speed as the nominal trajectory for over 8 laps without crashing, whereas the nominal controller crashes during its third lap after poor tracking. Results are shown in \autoref{fig:drift_traj} and \autoref{fig:drift_params}.

Both the cornering-stiffness and friction estimates by \llampc\ converge, and the \llampc\ model achieves substantially lower prediction error than the nominal model.

\subsection{Experiment 2: Changing Surfaces}
\begin{figure}[th]
    \centering
    \vspace{-2mm}
    \includegraphics[width=.49\linewidth]{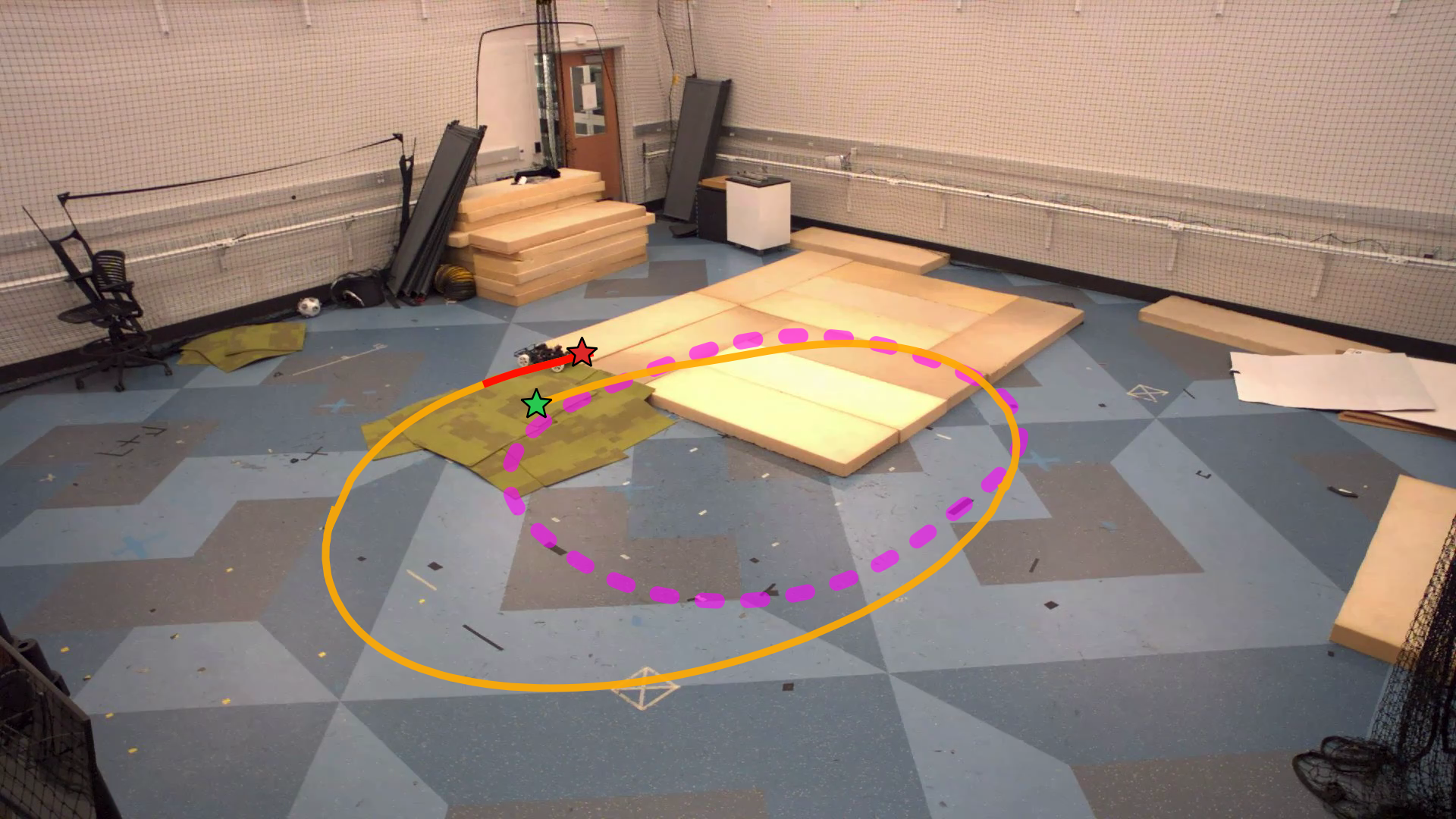}
    \includegraphics[width=.49\linewidth]{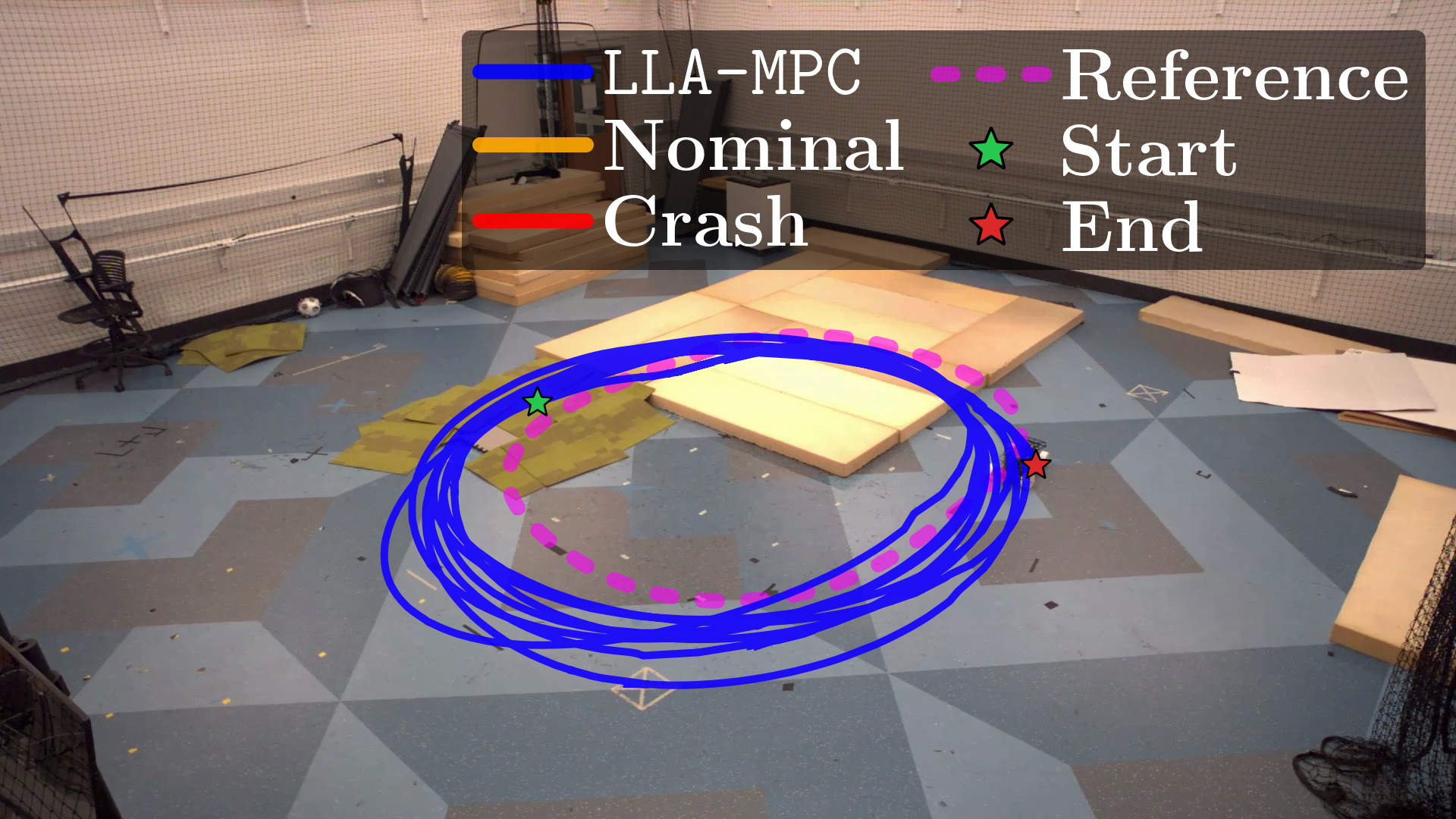}
    \caption{Nominal model tracking shown to the left. \llampc\ shown on the right.}
    \label{fig:jump_traj}
    \vspace{-3mm}
\end{figure}

\begin{figure}[th]
    \centering
    \vspace{-2mm}
    \includegraphics[width=0.49\linewidth]{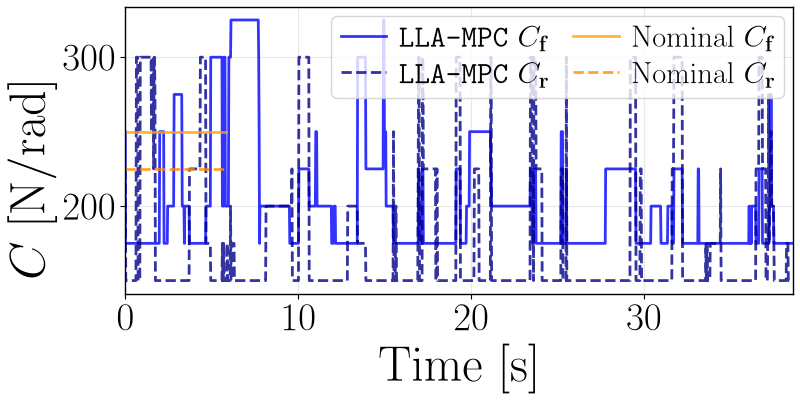}
    \includegraphics[width=.49\linewidth]{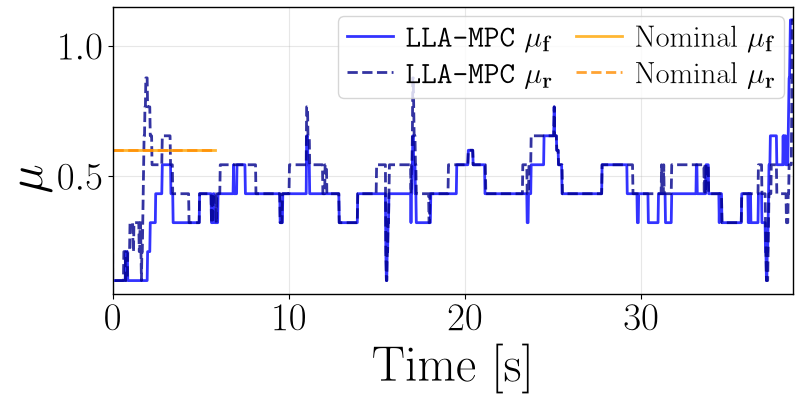}
    \includegraphics[width=.49\linewidth]{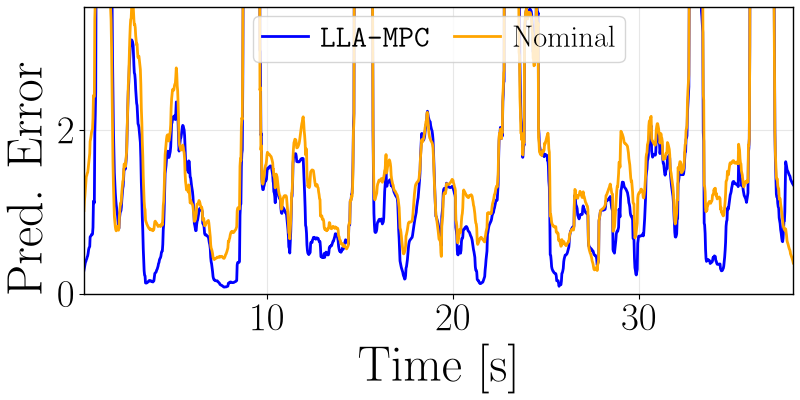}
    \includegraphics[width=.49\linewidth]{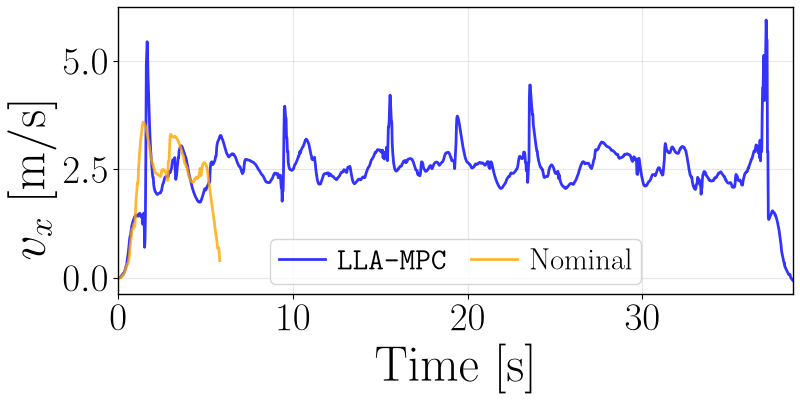}
    \caption{Model parameter estimation, longitudinal velocity, and $H$-step prediction error for run with \llampc. }
    \label{fig:jump_params}
\end{figure}

In the next experiment, the car tracks the same trajectory, but now across a variety of different surfaces and with an unmodeled abrupt surface height change. \llampc\ is able to command the car through over 6 laps successfully, whereas the nominal controller is only capable of completing a single lap due to lateral drift induced by high slippage inciting a crash, shown in \autoref{fig:jump_traj}. \llampc\ also shows adaptation between high-friction and low-friction surfaces, shown in \autoref{fig:jump_params}. $\mu$ is estimated lower when on the low-friction ground and higher when on higher-grip foam.  Predictions by \llampc\ had lower average error than predictions made by the nominal model over the successful \llampc\ run.

\subsection{Experiment 3: Onboard Compute}
\begin{figure}[!tp]
    \centering
    \vspace{-2mm}

    \includegraphics[width=\linewidth]{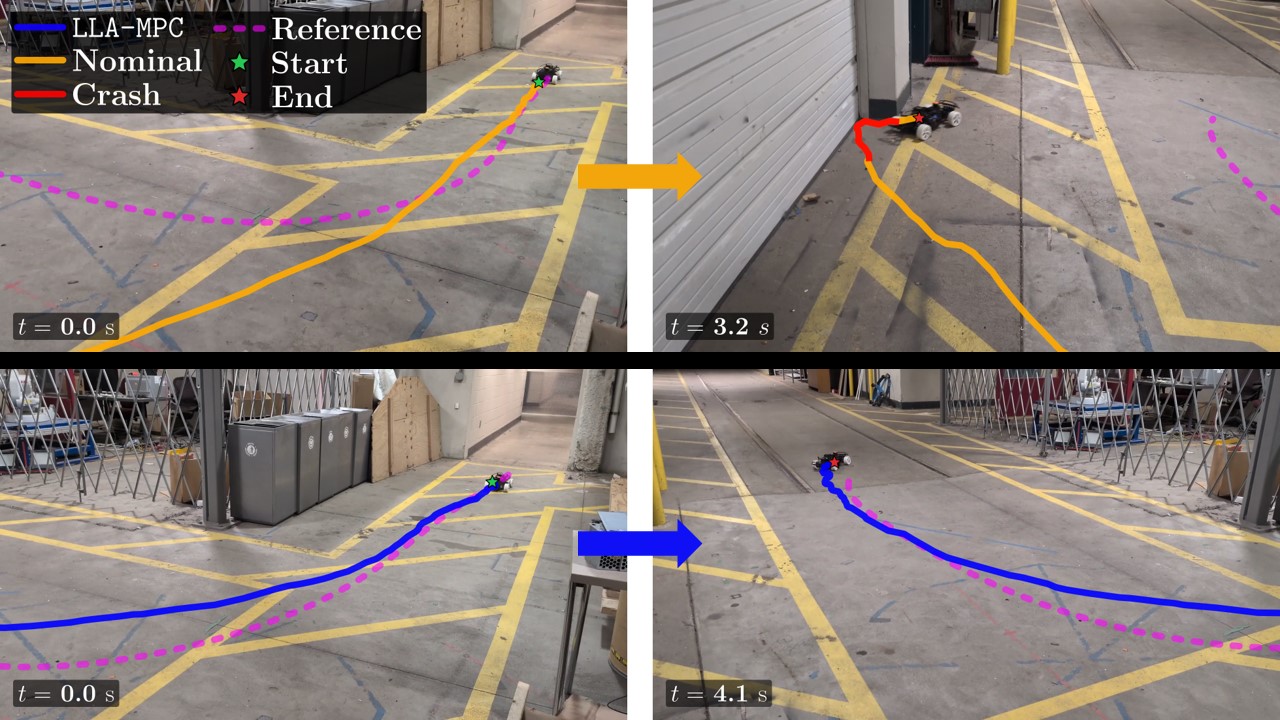}

    \caption{Nominal model tracking shown on the top two images. \llampc\ shown on the bottom two.}
    \label{fig:blevel_turn_traj}
\end{figure}

\begin{figure}[!tp]
    \centering
    \vspace{-2mm}
    \includegraphics[width=0.49\linewidth]{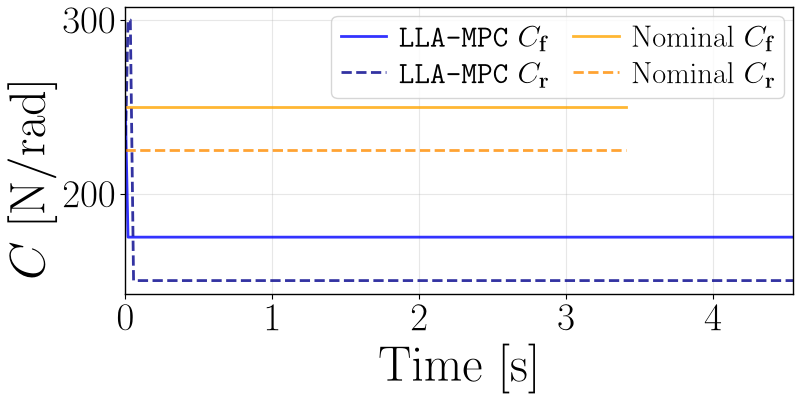}
    \includegraphics[width=.49\linewidth]{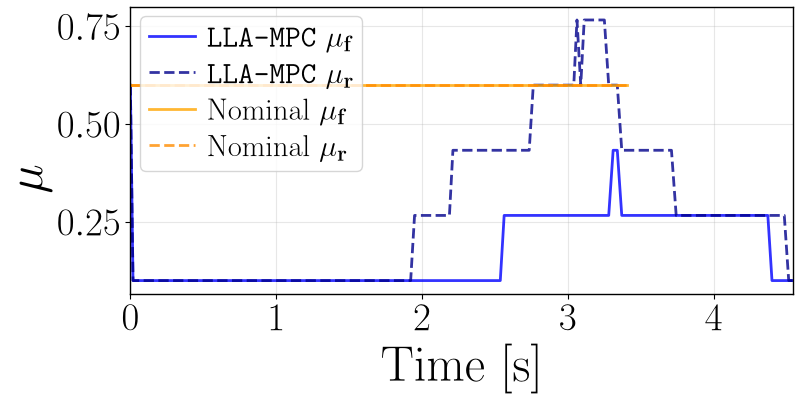}
    \includegraphics[width=.49\linewidth]{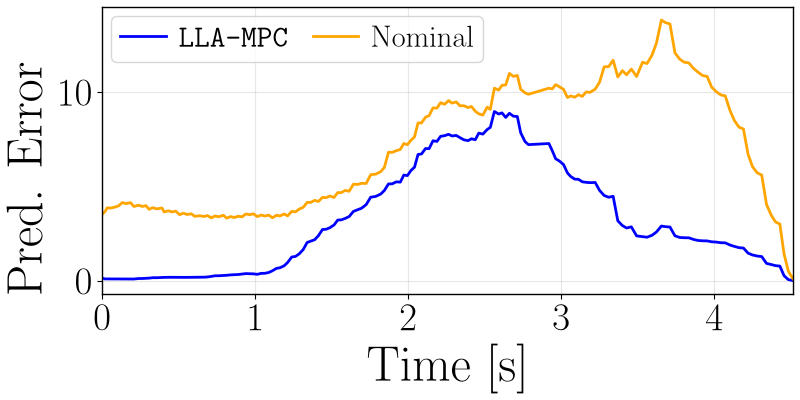}
    \includegraphics[width=.49\linewidth]{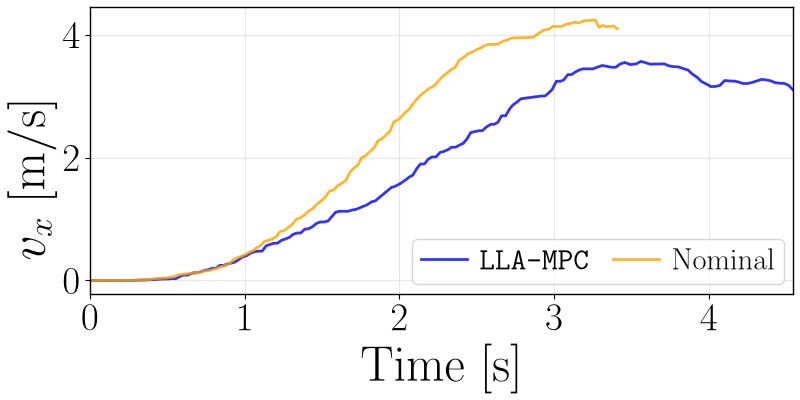}
    \caption{Model parameter estimation, longitudinal velocity, and $H$-step prediction error for run with \llampc. }
    \label{fig:blevel_turn_params}
    \vspace{-5mm}
\end{figure}

We also deploy the \llampc\ framework alongside noisy state estimation onboard the F1TENTH system, and command the car to attempt a fast turn on an extremely low-friction surface. The nominal model tracks the trajectory poorly and crashes the car  after a wide arc deviating far from the reference trajectory, whereas the \llampc\ controller is capable of accomplishing the drifting maneuver at high speed as shown in \autoref{fig:blevel_turn_traj} despite the adverse environment, high initial tire slip, and noisy state estimation. \autoref{fig:blevel_turn_params} shows that cornering stiffness estimates converged at the lower bound of the model bank, and \llampc\ outperformed the nominal model in $H$-step prediction error over the entire successful run.

\section{Future Work}
\label{sec:conclusion}

\llampc\ has potential areas for improvement that are yet to be explored. These include: weighting timesteps in the {\it look-back} cost calculation based on uncertainty in state estimation and model excitation, surveying \llampc\ performance with different sets of hyperparameters such as window size $W$, state weights $S$ and bank size, and the integration of Look-Back adaptive system identification with not only model-based control methods, but also adaptive model-based planning, safety, and learning methods.

\printbibliography

\end{document}